\documentclass[10pt,twocolumn,letterpaper]{article}

\usepackage{cvpr}
\usepackage{times}
\usepackage{epsfig}
\usepackage{graphicx}
\usepackage{amsmath}
\usepackage{amssymb}

\usepackage{algorithm}
\usepackage{algorithmic}
\usepackage{caption}
\usepackage{subfigure}
\usepackage{multirow}
\usepackage{array}

\newcolumntype{L}[1]{>{\raggedright\let\newline\\\arraybackslash\hspace{0pt}}m{#1}}
\newcolumntype{C}[1]{>{\centering\let\newline\\\arraybackslash\hspace{0pt}}m{#1}}
\newcolumntype{R}[1]{>{\raggedleft\let\newline\\\arraybackslash\hspace{0pt}}m{#1}}

\usepackage[pagebackref=true,breaklinks=true,letterpaper=true,colorlinks,bookmarks=false]{hyperref}

\makeatletter
\def\hlinewd#1{%
  \noalign{\ifnum0=`}\fi\hrule \@height #1 \futurelet
   \reserved@a\@xhline}
\makeatother

\cvprfinalcopy 

\def\cvprPaperID{93} 

\ifcvprfinal\fi
\begin{document}

\title{Robust Multi-body Feature Tracker: A Segmentation-free Approach}

\author{Pan Ji$^{1}$, Hongdong Li$^{1}$, Mathieu Salzmann$^{2}$, and Yiran Zhong$^{1}$\\
$^{1}$ANU, Canberra; $^{2}$EPFL, Switzerland
}

\maketitle

\begin{abstract}
Feature tracking is a fundamental problem in computer vision, with applications in many computer vision tasks, such as visual SLAM and action recognition.
 This paper introduces a novel multi-body feature tracker that exploits a multi-body rigidity assumption to improve tracking robustness under a general perspective camera model. A conventional approach to addressing this problem would consist of alternating between solving two subtasks: motion segmentation and feature tracking under rigidity constraints for each segment. This approach, however, requires knowing the number of motions, as well as assigning points to motion groups, which is typically sensitive to the motion estimates. By contrast, here,
 we introduce a segmentation-free solution to multi-body feature tracking that bypasses the motion assignment step and reduces to solving a series of subproblems with closed-form solutions. Our experiments demonstrate the benefits of our approach in terms of tracking accuracy and robustness to noise.
\end{abstract}

\section{Introduction}

Feature tracking is a prerequisite for many computer vision tasks, such as visual SLAM and action recognition. 
Among all the feature tracking methods, the Kanade-Lucas-Tomasi (KLT) tracker~\cite{lucas1981iterative,tomasi1991detection,shi1994good}, although developed 30 years ago,
 still remains one of the most widely used techniques. 
One of the reasons for this popularity is its computational efficiency; the KLT tracker is local, in the sense that it treats each local region independently of the others, which makes it highly parallelizable. This locality, however, comes at a cost in tracking robustness: the tracking of each feature cannot benefit from intrinsic scene constraints, and thus often suffers from drift. 

The real-world scenes, however, are often strongly constrained. For example, in autonomous driving, most of the moving objects (cars, vehicles, pedestrian) are rigid, or quasi-rigid if seen from afar. Several methods have therefore been proposed to exploit this scene rigidity to improve feature tracking~\cite{torresani2002space,buchanan2007combining,poling2014better}. Unfortunately, these methods all assume an affine camera model and are thus ill-suited to handle strong perspective effects. More importantly, they work either as a post-processing step on an entire sequence~\cite{torresani2002space}, which is sensitive to initial tracking results and does not apply to online feature tracking, or within a temporal sliding window~\cite{buchanan2007combining,poling2014better}, which is sensitive to initialization in the first few frames. 

By contrast, in this paper, we introduce a novel feature tracker that takes advantage of {\em multi-body scene rigidity} to improve tracking robustness under a general {\em perspective camera model}.  A conventional approach to addressing this problem would consist of alternating between two subtasks: motion segmentation and feature tracking under rigidity constraints for each segment. This, however, suffers from the following drawbacks: First, it requires knowing the number of observed motions; and, second, it relies on assigning points to individual motions, which is very sensitive to the initial motion estimates.


Here, we introduce a {\em segmentation-free} multi-body feature tracker that overcomes these drawbacks. Specifically, our approach bypasses the motion assignment step by making use of subspace constraints derived directly from the epipolar constraints of multiple motions. As a result, our algorithm does not require prior knowledge of the number of motions. Furthermore, this allows us to formulate tracking as an optimization problem whose subproblems all have closed-form solutions.


We demonstrate the effectiveness of our method on both feature point tracking and frame-by-frame motion segmentation on real world sequences. Our experiments show that, by incorporating multi-motion constraints, our tracker yields better accuracies and is more robust to noise than the standard KLT tracker and the state-of-the-art tracking algorithm of~\cite{poling2014better}.

\section{Related Work}

The KLT tracker~\cite{tomasi1991detection,shi1994good} was derived from the Lucas-Kanade algorithm for image alignment~\cite{lucas1981iterative}. Feature tracking was achieved by optimizing the sum of squared differences between a template patch and an image patch with the Gauss-Newton method. It was later extended to handle relatively large displacements by the use of image pyramids~\cite{bouguet2001pyramidal}.



Global rigidity constraints have been incorporated in feature point tracking to improve robustness. For instance, Torresani and Bregler~\cite{torresani2002space} proposed to regularize tracking with a global low-rank constraint on the trajectory matrix of the whole sequence. They relied on the original KLT tracker to get a set of reliable tracks, and explicitly factorized the reliable trajectory matrix into two low-rank matrices with the rank given {\it a priori}. One of the low-rank matrices, called the motion parameter matrix, was then used to rectify the unreliable tracks. In short, this method can be viewed as a post-processing step on the results of the KLT tracker, and is therefore not suitable for online frame-to-frame tracking.

Instead of using the whole sequence, low-rank constraints~\cite{buchanan2007combining} and similar subspace priors~\cite{poling2014better} were applied within a temporal sliding window. Specifically, Buchanan and Fitzgibbon~\cite{buchanan2007combining} exploited the low-rank constraints within a Bayesian tracking framework, making predictions of the new location of a particular point using a low rank approximation obtained from the previous frames. Recently, Poling \etal~\cite{poling2014better} proposed a better feature tracker by adding soft subspace constraints to the original KLT tracker and jointly solving for the displacement vectors of all feature points.
These methods, however, assume an affine camera model within a temporal window, and are therefore ill-suited to handle strong perspective effects. Moreover, since the low-rank constraints are enforced in a temporal sliding window, these methods are sensitive to initialization in the first few frames.


By contrast,~\cite{piccini2014good} exploits perspective projection by making use of epipolar constraints to track edgels in two consecutive frames. This method, however, was specifically designed to model a single motion, and thus does not easily extend to the multi-body case. 

In the closely related optical flow literature, several methods have been devoted to improving robustness via rigidity constraints. For instance, Valgaerts \etal~\cite{valgaerts2008variational} introduced a variational model to jointly recover the fundamental matrix and the optical flow; Wedel \etal~\cite{wedel2008duality,wedel2009structure} leveraged the fundamental matrix prior as an additional weak prior within a variational framework. These methods, however, assume that the scene is mostly stationary (and thus a single fundamental matrix is estimated), and treat the dynamic parts as outliers~\cite{wedel2009structure}. Garg \etal~\cite{garg2011dense,garg2013variational} proposed to make use of subspace constraints to regularize the multi-frame optical flow within a variational approach. This approach, however, assumes an affine camera model and works over entire sequences. 

While, to the best of our knowledge, explicitly modeling multi-body motion has not been investigated in the context of feature tracking and optical flow estimation, a large body of work~\cite{costeira1998multibody, vidal2005generalized,yan2006general, li2007two,li2013perspective,vidal2002segmentation,elhamifar2013sparse,ji2014efficient,ji2014null,ji2015shape} has been devoted to multi-body motion segmentation given good point trajectories in relatively long sequences. Typically, these tracks are first obtained with the KLT tracker, and then manually cleaned up, \eg, the Hopkins155 dataset~\cite{tron2007benchmark}. In a sense, the lack of better tracking algorithms that can incorporate the intrinsic constraints of dynamic scenes prevents the practical use of these motion segmentation algorithms.

In this paper, we seek to track feature points in dynamic scenes where multiple motions are present. In this scenario, a single fundamental matrix is not sufficient to express the epipolar constraints any more. While one could think of alternating between estimating multiple fundamental matrices, motion assignments and displacement vectors, the resulting algorithm would typically be very sensitive to initialization, since the motion assignments strongly depend on the motion estimates. By contrast, we introduce a segmentation-free approach that bypasses the motion assignment problem by exploiting subspace constraints derived from epipolar geometry. This yields a robust multi-body tracking algorithm that, as demonstrated by our experiments, opens up the possibility to perform motion segmentation in realistic scenarios.


\section{Multi-body Feature Tracker}

We now introduce our approach to multi-body feature tracking. 
Formally, let $I(\bf x)$ denote the current image, $T(\bf x)$ the previous image (or template image), and ${\bf x}_{ij} = [x_{ij}, y_{ij}]^T$ the $j^{\rm th}$ image point in the $i^{\rm th}$ patch $\Omega_i$ of the template image. Our goal is to estimate the displacement vector ${\bf u} = [{\bf u}_i^T, \cdots, {\bf u}_N^T]^T \in \mathbb{R}^{2\times N}$ for all $N$ tracked feature points. To this end, we rely on the standard brightness constancy assumption~\cite{szeliski2010computer}, which lets us derive the data term
\begin{equation}
\mathcal{D}({\bf u}) = \sum\limits_{i=1}^N\sum\limits_{{\bf x}_{ij}\in\Omega_i}\psi\big(I({\bf x}_{ij}+{\bf u}_i) - T({\bf x}_{ij})\big)\;,
\end{equation}
where, typically, $\psi(x) = x^2$ or $\psi(x) = |x|$. In particular, we use the $\ell_1$ norm, which provides robustness to outliers.

Estimating the displacements from this data term only is typically sensitive to noise and may be subject to drift. A general approach to making the process more robust consists of introducing a regularizer $\mathcal{R}({\bf u})$ to form an energy function of the form
\begin{equation}
\mathcal{F}({\bf u}) = \gamma \mathcal{D}({\bf u}) + \mathcal{R}({\bf u})\;.
\end{equation}
As mentioned above, several attempts at designing such a regularizer have been proposed. For example, under an affine camera model, $\mathcal{R}({\bf u})$ can encode a low-rank prior~\cite{torresani2002space,poling2014better}; with a general projective camera model, $\mathcal{R}({\bf u})$ can represent epipolar constraints (\ie, a fundamental matrix prior)~\cite{wedel2008duality,valgaerts2008variational,piccini2014good}. In the latter case, the fundamental matrix can be either pre-computed via an existing feature matching method~\cite{piccini2014good}, or re-computed iteratively.

When multiple motions are present, however, a single epipolar constraint is not sufficient. Instead, multiple fundamental matrices should be estimated so as to respect the assignments of the tracked points to individual motions. A straightforward way to addressing this problem consists of adding a motion segmentation step in the tracking algorithm, so that the fundamental matrices can be iteratively re-estimated. This leads to the simple {\em segmentation-based} approach to multi-body feature tracking as described below.


\subsection{A First Attempt: Segmentation-based Tracking}
\label{subsec:seg-based}

To derive a segmentation-based approach, we rely on epipolar constraints. Recall that, in epipolar geometry~\cite{Hartley2004}, the homogeneous coordinates $\bar{\bf x}_i^{\prime} = ({x}_i^{\prime},{y}_i^{\prime},1)^T$ and $\bar{\bf x}_i = ({x}_i,{y}_i,1)^T$ of two corresponding image points in two frames are related by a fundamental matrix ${\bf F}$, such that
\begin{equation}
\bar{\bf x}_i^{\prime T} {\bf F}\bar{\bf x}_i = 0\;.
\label{eq:fundamental}
\end{equation}
It is therefore natural to exploit these constraints to regularize tracking according to the motion assignments of the different points.

More specifically, in the segmentation-based approach, three types of variables must be estimated: the displacement vector ${\bf u}$, the fundamental matrices $\{{\bf F}^k\}_{k = 1,\cdots,K}$ (where $K$ is the number of motions), and the motion label of each tracked point. Let us denote by $\bar{\bf x}_i^k$ the homogeneous coordinate of the $i^{\rm th}$ feature point (\ie, the center of the patch $\Omega_i$) assigned to motion $k$. We can define a multi-body regularization term as
\begin{equation}
\label{eq:seg_based}
\mathcal{R}_1({\bf u},{\bf F}^k) = \sum\limits_{k}\sum\limits_{i}\big[(\bar{\bf x}_i^k+\bar{\bf u}_i)^T{\bf F}^k\bar{\bf x}_i^k\big]^2\;,
\end{equation}
where $\bar{\bf u}_i = [{\bf u}_i^T,0]^T$.

The energy function can then be approximately minimized by iterating over the three following steps:
\begin{enumerate}
\setlength{\itemsep}{0pt}
\setlength{\parskip}{0pt}
\item Update ${\bf u}$ by first-order gradient descent~\cite{poling2014better};
\item Estimate ${\bf F}^k$ for each motion given the current point assignments;
\item Re-assign the motion labels of the feature points to the nearest ${\bf F}^k$.
\end{enumerate}

This segmentation-based approach suffers from several drawbacks. First, the number of motions needs to be known {\it a priori}, which is typical hard for general-purpose tracking. Second, and more importantly, the quality of solution obtained with this approach will strongly depend on the initializations of ${\bf F}^k$ and of the motion labels. This, in a sense, is a chicken-and-egg problem, since good initialization for these variables could be obtained from good motion estimates. 
Instead, in the remainder of this section, we introduce a new segmentation-free approach that bypasses the need to explicitly compute the fundamental matrices and the motion assignments.


\subsection{ Our Segmentation-free Approach}
\label{subsec:seg-free}

In this section, we introduce our segmentation-free multi-body feature tracker, which is the key contribution of this paper. We first show how the epipolar constraints can be converted to subspace constraints, and incorporated into our tracking formalism. We then derive the solution to the resulting optimization problem by decomposing it into several convex subproblems all with closed-form solutions.

\subsubsection{Epipolar Subspace Constraints}

As in the segmentation-based approach, we seek to rely on epipolar geometry. To this end, we make use of the constraint expressed in Eq.~\ref{eq:fundamental}. We first note that this constraint can be re-writen as
\begin{equation}
\label{eq:fundm}
{\bf f}^T {\rm vec}(\bar{\bf x}_i^{\prime} \bar{\bf x}_i^T) = 0\;,
\end{equation}
where ${\bf f}\in\mathbb{R}^9$ is the vectorized fundamental matrix ${\bf F}$, and 
\begin{equation}
{\rm vec}(\bar{\bf x}_i^{\prime} \bar{\bf x}_i^T) = (x_i{x}_i^{\prime},x_i{y}_i^{\prime},x_i,y_i{x}_i^{\prime},y_i{y}_i^{\prime},y_i,{x}_i^{\prime},{y}_i^{\prime},1)^T\;.
\end{equation}
Let us define ${\bf w}_i = {\rm vec}(\bar{\bf x}_i^{\prime} \bar{\bf x}_i^T)$. Then, ${\bf w}_i$ lies in the orthogonal complement of ${\bf f}^T$, which is a subspace of dimension up to eight
\footnote{Note that, in practice, this dimension is typically smaller than 8, since, in real scenes, the motion of objects, such as cars or people, is not arbitrary, and thus corresponds to degenerate (\ie, low-rank) motion~\cite{li2013perspective}.}, and which we call the {\em epipolar subspace}. Since image points undergoing the same motion share the same fundamental matrix, all ${\bf w}_i$s corresponding to points belonging to the same rigid motion lie on the same subspace~\cite{li2013perspective}.

Therefore, in our multi-body feature tracking scenario, if the feature points are correctly tracked, the data vectors defined as
\begin{equation}
\label{eq: w_i}
{\bf w}_i = {\rm vec}\big((\bar{\bf x}_i+\bar{\bf u}_i) \bar{\bf x}_i^T\big) \;, \;\;\;\;\forall\; 1 \leq i \leq N\;,
\end{equation}
should lie in a union of linear subspaces. This subspace constraint can be characterized by the self-expressiveness property~\cite{elhamifar2013sparse,ji2014efficient}, \ie, a data point drawn from one subspace in a union of subspaces can be represented as a linear combination of the points lying in the same subspace.

In our case, this self-expressiveness property can be expressed as
\begin{equation}
{\bf W}_{(\bf u)} = {\bf W}_{(\bf u)}{\bf C}\;,
\label{eq:self_expr}
\end{equation}
where ${\bf W}_{(\bf u)} = [{\bf w}_1\cdots{\bf w}_N]$\footnote{In the following, we make use of subscript $({\bf u})$, \ie, ${\bf W}_{({\bf u})}$, to indicate that ${\bf W}$ depends on the variable ${\bf u}$. For compactness, and without causing confusion, we drop this explicit dependency in Section~\ref{sec:admm}.}, and ${\bf C}$ is the coefficient matrix encoding the linear combinations. On its own, this term has a trivial solution for ${\bf C}$ (\ie, the identity matrix). To avoid this solution, ${\bf C}$ needs to be regularized. In the subspace clustering literature, ${\bf C}$ is encouraged to be either sparse~\cite{elhamifar2013sparse} by minimizing $\|{\bf C}\|_1$, low rank~\cite{liu2013robust} by minimizing $\|{\bf C}\|_*$, or dense block diagonal~\cite{ji2014efficient} by minimizing $\|{\bf C}\|_F^2$. Here, we choose the Frobenius norm, which has proven effective and is easy to optimize. Furthermore, we explicitly model noise and outliers, which are inevitable in real-world sequences. 

More specifically, we write our regularization term for multi-body tracking as
\begin{equation}
\label{eq:subspace_reg}
\mathcal{R}_2({\bf u},{\bf C}) = \frac{1}{2}\|{\bf C}\|_F^2+\lambda\|{\bf E}\|_1\;,\;{\rm s.t.}\; {\bf W}_{(\bf u)} = {\bf W}_{(\bf u)}{\bf C}+ {\bf E}\;,
\end{equation}
where ${\bf E}$ accounts for noise and outliers, and is thus encouraged to be sparse. Note that, for a given displacement $\bf u$, and ignoring noise, the optimal value of this regularizer depends on the intrinsic dimension of the motion~\cite{ji2014efficient}. Since here we optimize ${\bf u}$, this regularizer therefore tends to favor degenerate rigid motions over purely arbitrary rigid motions. This actually reflects reality, since, in real scenes, cars, people and other objects typically move in a well-constrained manner.

Importantly, this regularization term requires explicitly computing neither the fundamental matrices, nor the motion assignments. As such, it therefore yields a {\em segmentation-free} approach.

Altogether, the energy function of our multi-body tracking framework can be written as
\begin{equation}
\mathcal{F}({\bf u},{\bf C}) = \gamma \mathcal{D}({\bf u}) + \mathcal{R}_2({\bf u},{\bf C})\;.
\label{eq:full_obj}
\end{equation}
Our goal is to minimize $\mathcal{F}({\bf u},{\bf C})$ w.r.t. ${\bf u}$ and ${\bf C}$. We next show how to solve this optimization problem.

\subsubsection{Approximation and Problem Reformulation}

To optimize Eq.~\ref{eq:full_obj}, we first approximate the data term in the same manner as the original KLT. In other words, given an initial displacement ${\bf u}_i^0$ for patch $i$, we approximate the intensity values $I({\bf x}_{ij} + {\bf u}_i)$
with their first-order Taylor expansion at ${\bf x}_{ij}+{\bf u}_i^0$. This can be written as
\begin{equation}
\label{eq:taylor_approx}
I({\bf x}_{ij}+{\bf u}_i) \approx I({\bf x}_{ij}+{\bf u}_i^0) + \triangledown I({\bf x}_{ij}+{\bf u}_i^0)({\bf u}_i-{\bf u}_i^0)\;.
\end{equation}
For notational convenience, let $\triangledown I_{ij} = \triangledown I({\bf x}_{ij}+{\bf u}_i^0)$, and $\tau_{ij} = \triangledown I_{ij}{\bf u}_i^0+T({\bf x}_{ij})-I({\bf x}_{ij}+{\bf u}_i^0)$. Then, the data term can be expressed as
\begin{equation}
\mathcal{D}({\bf u}) = \sum\limits_{i,j}|\triangledown I_{ij}{\bf u}_i-\tau_{ij}|\;.
\end{equation}
By combining this data term with our regularizer, we get the optimization problem
\begin{equation}
\begin{split}
\min\limits_{{\bf u},{\bf C},{\bf E}} \gamma \|{\bf A}_{(\bf u)}\|_1 + \frac{1}{2}\|{\bf C}\|_F^2 + \lambda\|{\bf E}\|_1\hspace{0.0cm}\\
{\rm s.t.}\; {\bf W}_{(\bf u)} = {\bf W}_{(\bf u)}{\bf C}+ {\bf E}\;,\hspace{0.8cm}
\label{eq:prob_original}
\end{split}
\end{equation}
where ${\bf A}_{ij} = \triangledown I_{ij}{\bf u}_i-\tau_{ij}$.

For convenience of optimization, we introduce an auxiliary variable ${\bf Z} = {\bf A}_{(\bf u)}$. Then,~\eqref{eq:prob_original} can be equivalently written as
\begin{equation}
\label{eq:optZ}
\begin{split}
\min\limits_{{\bf u},{\bf C},{\bf E},{\bf Z}} \gamma\|{\bf Z}\|_1 + \frac{1}{2}\|{\bf C}\|_F^2 + \lambda\|{\bf E}\|_1\hspace{0.5cm}\\
{\rm s.t.}\quad {\bf Z} = {\bf A}_{(\bf u)}\;,{\bf W}_{(\bf u)} = {\bf W}_{(\bf u)}{\bf C}+ {\bf E}\;. \hspace{0cm}
\end{split}
\end{equation}
The main hurdle in optimizing~\eqref{eq:optZ} now lies in the term with ${\bf W}_{(\bf u)}$ due to its seemingly complicated dependency on ${\bf u}$. However, we show below that this term can be simplified by a few matrix derivations.

First, note that, by definition, we have
\begin{equation}
\label{defP}
{\rm vec}({\bf W}_{(\bf u)}) = \underbrace{\begin{bmatrix} \bar{\bf x}_1 \otimes {\bf I}_{3\times 3} & & \\ &\ddots& \\ & & \bar{\bf x}_N \otimes {\bf I}_{3\times 3} \end{bmatrix}}_{\bar{\bf P}}(\bar{\bf x}+\bar{\bf u})\;,
\end{equation}
where $\bar{\bf x} = [\bar{\bf x}_1^T \cdots \bar{\bf x}_N^T]^T$, $\bar{\bf u} = [\bar{\bf u}_1^T \cdots \bar{\bf u}_N^T]^T$, ${\bf I}_{3\times 3}$ is the 3-by-3 identity matrix and $\otimes$ denotes the Kronecker product.
Let us define ${\bf b} = \bar{\bf P}\bar{\bf x}$ (or equivalently ${\bf b}_i = {\bf vec}(\bar{\bf x}_i \bar{\bf x}_i^T)$ ) and introduce another auxiliary variable ${\bf m} = {\bf P}{\bf u}$ (where ${\bf P}$ is obtained by removing every ${3i}^{th}$ column of $\bar{\bf P}$)\footnote{Note that ${\bf P}{\bf u}$ = $\bar{{\bf P}}\bar{{\bf u}}$, since $\bar{\bf u}_i = [{\bf u}_i^T,0]^T$.}. Our optimization problem then becomes
\begin{equation}
\label{eq:optZm}
\begin{split}
\min\limits_{{\bf u},{\bf C},{\bf E},{\bf Z},{\bf m}} \gamma\|{\bf Z}\|_1 + \frac{1}{2}\|{\bf C}\|_F^2 + \lambda\|{\bf E}\|_1\hspace{1cm}\\
{\rm s.t.}\quad {\bf Z} = {\bf A}_{(\bf u)}\;,{\bf W}_{(\bf m)} = {\bf W}_{(\bf m)}{\bf C}+ {\bf E}\;, {\bf m} = {\bf P}{\bf u}\;,
\end{split}
\end{equation}
where now ${\rm vec}({\bf W}_{(\bf m)}) = {\bf b} + {\bf m}$.

The above optimization problem involves a large number of variables.
We propose to solve it via the Alternating Direction Method of Multipliers (ADMM)~\cite{boyd2011distributed}, which decomposes a big optimization problem into several small subproblems. Below, we show how this can be achieved for our problem.

\subsubsection{ADMM Solution}
\label{sec:admm}
\begin{algorithm}[t!]
\caption{Solving~\eqref{eq:optZm} via the ADMM}
\label{Algorithm 1}
\begin{algorithmic}
\REQUIRE ~~\\
Image $I$ and template $T$, positions of the feature points ${\bf x}$ in $T$, initial displacement vector ${\bf u}^0$, parameters $\gamma$, $\lambda$.
 \\ \vspace{0.2cm}
\hspace{-0.3cm}{\bf Initialize:} ${\bf C}$ = 0, ${\bf Y}_1 = 0$, ${\bf Y}_2 = 0$, ${\bf y} = 0$, ${\bf A}_{({\bf u}^0)}$, ${\bf W}_{({\bf u}^0)}$, $\rho_0$, $\rho_m$, $\eta$, $\epsilon$\\ \vspace{0.2cm}
\WHILE {not converged}
\STATE 1. Update ${\bf Z}$, ${\bf E}$, ${\bf C}$, ${\bf u}$ and ${\bf m}$ in close-form via Eqs.~\ref{eq:sol_Z}-~\ref{eq:sol_m}, respectively;

\STATE 2. Update ${\bf A}_{(\bf u)}$ and ${\bf W}_{(\bf m)}$ with updated ${\bf u}$ and ${\bf m}$;

\STATE 3. Update the Lagrange multipliers and penalty parameter via Eqs.~\ref{eq:updatey}-~\ref{eq:updaterho};

\STATE 4. Check the convergence conditions $\|{\bf m}-{\bf P}{\bf u}\|_{\infty} \leq \epsilon$, $\|{\bf W}_{(\bf m)}-{\bf W}_{(\bf m)}{\bf C}-{\bf E}\|_{\infty} \leq \epsilon$, and $\|{\bf Z}-{\bf A}_{(\bf u)}\|_{\infty} \leq \epsilon$; \\
\ENDWHILE
\vspace{0.2cm}
\ENSURE ~Displacement vector ${\bf u}$, coefficient matrix ${\bf C}$.
\end{algorithmic}
\end{algorithm}

To apply the ADMM, we first need to derive the augmented Lagrangian of~\eqref{eq:optZm}, which can be expressed as
{\small
\begin{eqnarray}
\mathcal{L}_{\rho} = \gamma\|{\bf Z}\|_1 + \frac{1}{2}\|{\bf C}\|_F^2 + \lambda\|{\bf E}\|_1 + {\bf y}^T({\bf m} - {\bf P}{\bf u}) +\hspace{0.5cm}\\ \langle{\bf Y}_1, {\bf W}-{\bf W}{\bf C}-{\bf E}\rangle+ \langle{\bf Y}_2,{\bf Z}-{\bf A}_{(\bf u)}\rangle +  \hspace{1cm}\nonumber\\
 \frac{\rho}{2}\big(\|{\bf W}-{\bf W}{\bf C}-{\bf E}\|_F^2+\|{\bf Z}-{\bf A}_{(\bf u)}\|_F^2 + \|{\bf m} - {\bf P}{\bf u}\|_2^2\big)\;, \nonumber
\label{eq:augm}
\end{eqnarray}
}where $\langle\cdot ,\cdot \rangle$ denotes the matrix inner product, ${\bf Y}_1$,${\bf Y}_2$, ${\bf y}$ are Lagrange multipliers, and $\rho$ is the penalty parameter.
The ADMM then works by alternatively minimizing $\mathcal{L}_{\rho}$ w.r.t. one of the five variables ${\bf u}$, ${\bf C}$, ${\bf E}$, ${\bf Z}$, ${\bf m}$ while keeping the remaining four fixed.

As shown in appendix, the five subproblems derived from the augmented Lagrangian are all convex problems that can be solved efficiently in closed-form.
These closed-form solutions can be written as
{\small
\begin{align}
\label{eq:sol_Z}
{\bf Z} &= \mathcal{T}_{\frac{\gamma}{\rho}}[{\bf A}_{(\bf u)} - {\bf Y}_2/\rho]\;,\\
\label{eq:sol_E}
{\bf E} &= \mathcal{T}_{\frac{\lambda}{\rho}}[{\bf W}-{\bf W}{\bf C} + {\bf Y}_1/\rho]\;,\\
\label{eq:sol_C}
{\bf C} &= ({\bf I}+\rho {\bf W}^T{\bf W})^{-1}[\rho {\bf W}^T({\bf W}-{\bf E}+{\bf Y}_1/\rho)]\;,\\
\label{eq:sol_u}
{\bf u} &= (\rho{\bf P}^T{\bf P}+\rho{\bf H})^{-1}({\bf g}+{\bf P}^T{\bf y}+\rho{\bf P}^T{\bf m})\;,\\
\label{eq:sol_m}
{\bf M} &= -(\rho{\bf G} + {\bf B}{\bf Q} + {\bf T})(\lambda{\bf Q}+\rho{\bf I})^{-1}\;,
\end{align}
}where ${\bf m} = {\rm vec}({\bf M})$ is the vectorized form of ${\bf M}$, $\mathcal{T}_{\alpha}[x] = {\rm sign}(x)\cdot \max(|x|-\alpha,0)$ is the soft-thresholding operator, and the definitions of ${\bf g}$, ${\bf H}$, ${\bf Q}$, ${\bf T}$ are given in appendix.

Finally, the Lagrange multipliers and penalty parameter can be updated as
{\small
\begin{align}
\label{eq:updatey}
{\bf Y}_1 &= {\bf Y}_1 + \rho({\bf W}-{\bf W}{\bf C}-{\bf E})\;,\\
{\bf Y}_2 &= {\bf Y}_2 + \rho({\bf Z}-{\bf A}_{(\bf u)})\;,\\
{\bf y}\;\;&= {\bf y}   + \rho({\bf m} - {\bf P}{\bf u})\;\\
\label{eq:updaterho}
\rho\;\;  &= \min(\eta\rho,\rho_m)\;,
\end{align}
}where $\eta > 1$, and $\rho_m$ is the predefined maximum of $\rho$.

Our approach to solving~\eqref{eq:optZm} is outlined in Algorithm~\ref{Algorithm 1}. Note that the problem we are trying to solve is non-convex in that i) the intensity function $I({\bf x};{\bf u})$ is non-convex w.r.t. ${\bf u}$; ii) the optimization problem~\ref{eq:optZm} involves a bilinear term in an equality constraint. While the ADMM does not guarantee convergence to the global optimum, it has proven effective in practice~\cite{ji2014robust}.

\subsubsection{Our Complete Multi-body Feature Tracker}

In the same spirits as~\cite{bouguet2001pyramidal}, we make use of an image pyramid to handle large displacements and avoid local optima. The results obtained at a coarser level $\ell$ of the pyramid are used as initialization for the next ($\ell-1$, finer) level. Within each pyramid level, the initial displacement ${\bf u}^0$ , where the first-order Taylor approximation is performed, is updated with the displacement vector of the previous iteration. We iterate over successive Taylor approximations until the displacement vector does not change significantly. Our complete segmentation-free multi-body feature tracker is outlined in Algorithm~\ref{Algorithm 2}.

\begin{algorithm}[t!]
\caption{Our Multi-body Feature Tracker}
\label{Algorithm 2}
\begin{algorithmic}
\REQUIRE ~~\\
Image $I$ and template $T$, positions of the feature points ${\bf x}$ in $T$, initial displacement vector ${\bf u}^0$, number of pyramid levels $L$, parameters $\gamma$, $\lambda$, $\rho$, $\rho_m$, ${\rm max}_i$, $\epsilon$
 \\ \vspace{0.2cm}
\FOR{$\ell = L-1:0$}
\STATE Update ${\bf u}^0 \leftarrow {\bf u}^0/2^{\ell}$, ${\bf x} \leftarrow {\bf x}^0/2^{\ell}$ and compute $\triangledown I$ at current image pyramid level;
\vspace{0.1cm}

\FOR{$i = 1:{\rm max}_i$}

\STATE 1. Approximate the image intensities with Eq.~\ref{eq:taylor_approx}, and compute $\tau$, ${\bf P}$, ${\bf H}$ according to their definitions;

\STATE 2. Update {\bf u} with Algorithm~\ref{Algorithm 1};

\STATE 3. Check the convergence condition $\|{\bf u}-{\bf u}^0\|<\epsilon$;

\STATE 4. If not converged, update ${\bf u}^0 = {\bf u}$.

\ENDFOR

\vspace{0.1cm}
\STATE Update ${\bf u} \leftarrow 2^{\ell}{\bf u}$, ${\bf u}^0 \leftarrow 2^{\ell}{\bf u}^0$, and ${\bf x} \leftarrow 2^{\ell}{\bf x}$.

\ENDFOR
\vspace{0.2cm}
\ENSURE ~Displacement vector ${\bf u}$, coefficient matrix ${\bf C}$.
\end{algorithmic}
\end{algorithm}


\section{Experiments}

\begin{figure*}[!t]
\centering
  \begin{tabular}{ccccc}
\hspace{-0.3cm}  \includegraphics[height=0.18\linewidth]{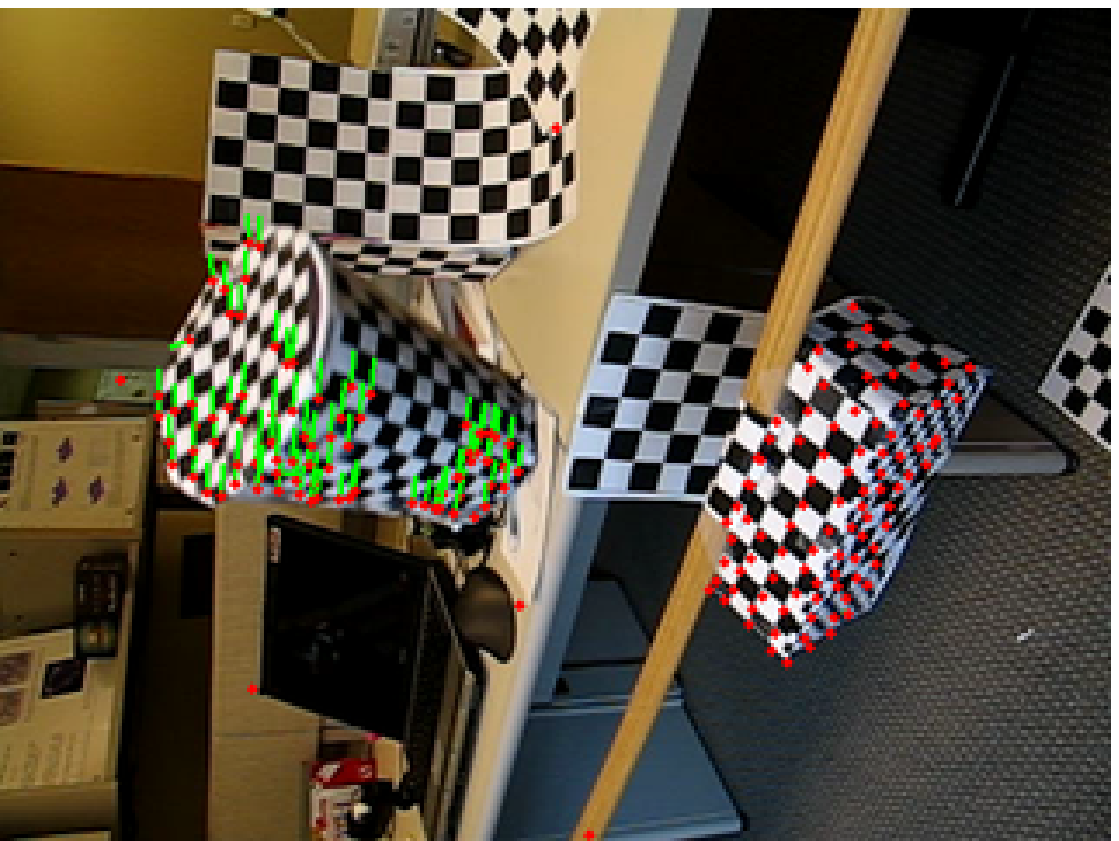} &
\hspace{-0.4cm}  \includegraphics[height=0.18\linewidth]{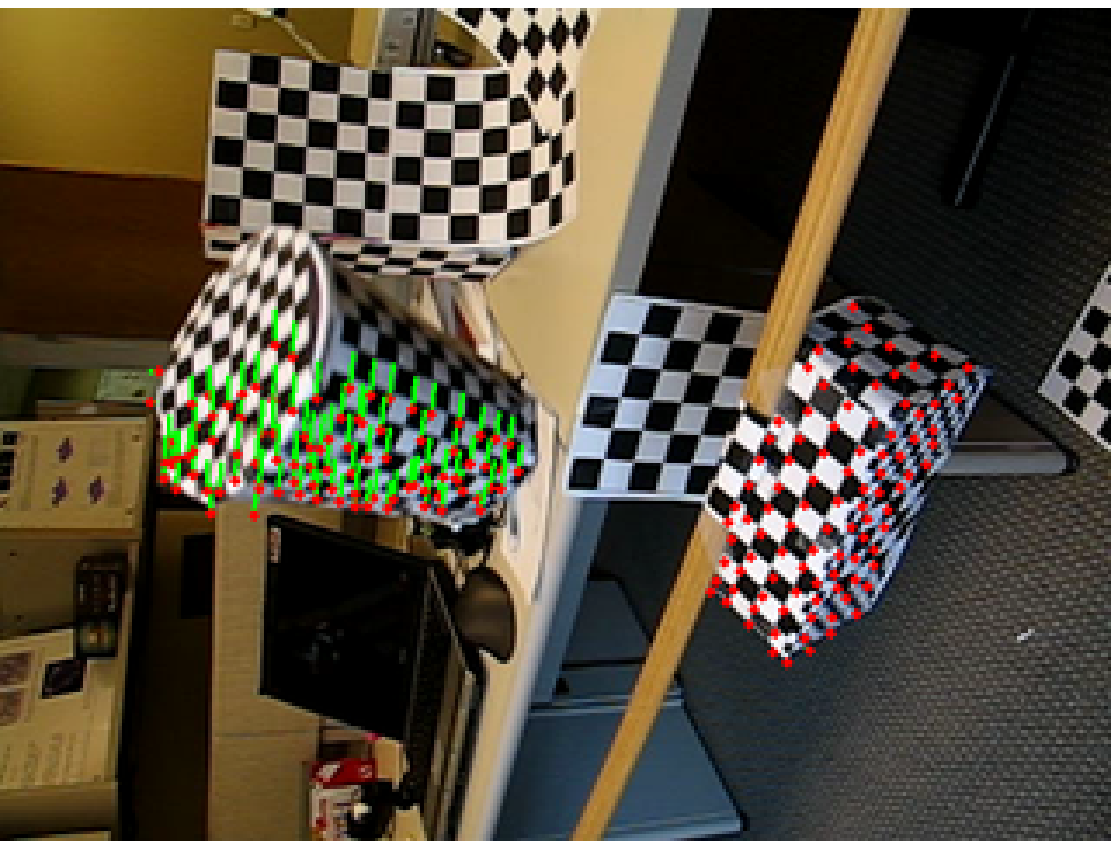} &
\hspace{-0.4cm}  \includegraphics[height=0.18\linewidth]{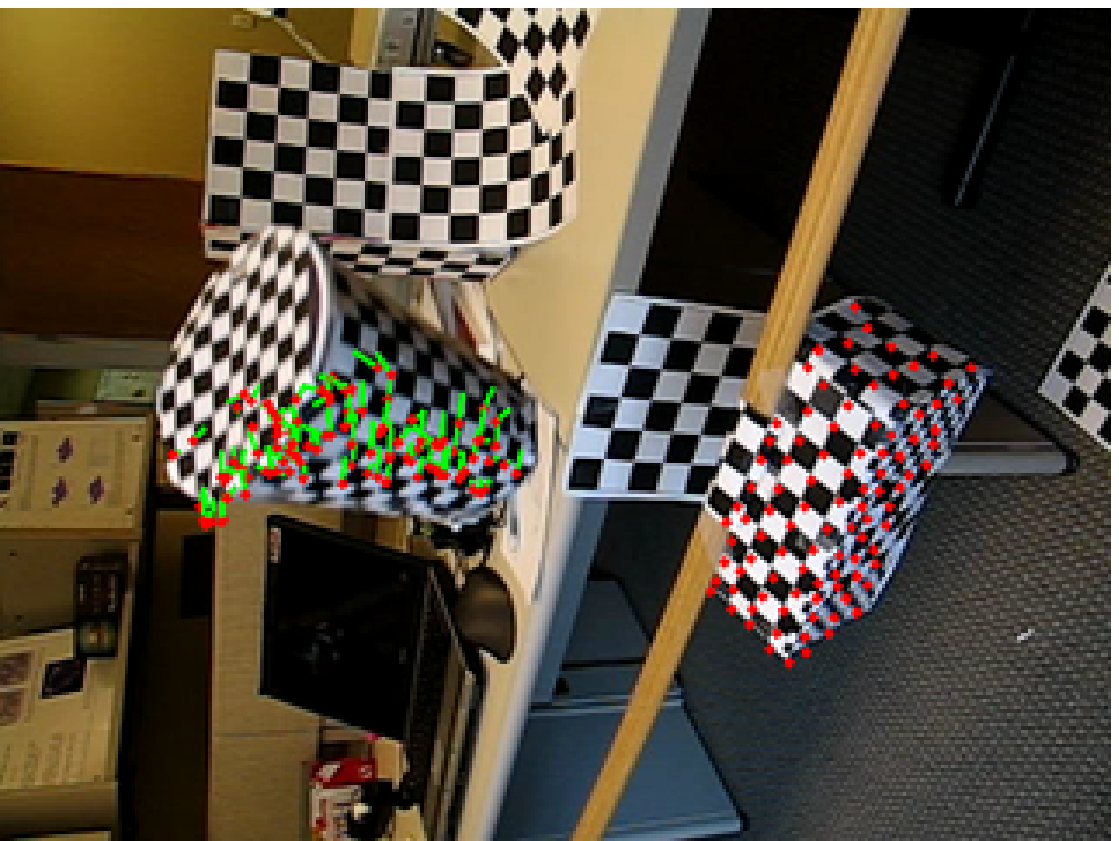} &
\hspace{-0.4cm}  \includegraphics[height=0.18\linewidth]{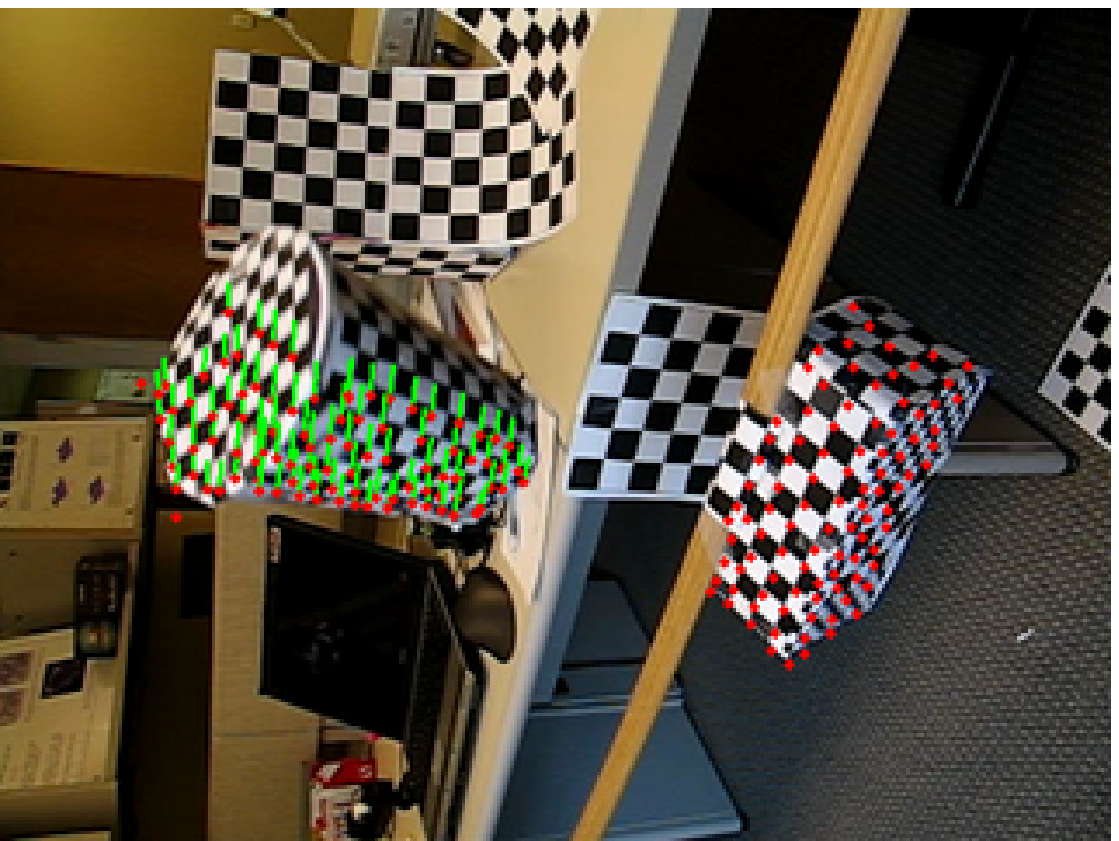}
\\
 \hspace{-0.3cm} KLT & \hspace{-0.4cm}L1-KLT & \hspace{-0.4cm}BFT & \hspace{-0.4cm}Our Method
  \end{tabular}
  \vspace{-0.3cm}
  \caption{{\bf Performance of different trackers on the 1RT2TC checkerboard sequence:} The red points denote the current positions of the feature points, and the green lines the motion since the previous frame. Best viewed on screen with zoom-in.}
  \label{fig:hopkins_checkerboard}
\end{figure*}
%

To show the benefits of our multi-body feature tracker, we performed extensive experiments on different sequences. In the remainder of this section, we present both qualitative and quantitative results.

In these experiments, we compare our approach with the following baselines: the original KLT tracker (KLT), the L1-norm KLT tracker (L1-KLT), and the more recent Better Feature Tracker (BFT) through Subspace Constraints~\cite{poling2014better}. For the original KLT, we used the Matlab built-in vision toolbox {\it vision.PointTracker}; we implemented the L1-norm KLT tracker using the same framework as our method by just disabling the regularization term; and for BFT, we used the code released by the authors.

Due to the lack of benchmark datasets for feature tracking, we make use of motion segmentation datasets where both the ground-truth tracks and the original videos are available. Since those videos are typically only provided for illustration purpose, they are generally highly compressed and not ideal for reliable feature tracking. This, however, is not really a problem when one seeks to evaluate feature tracking methods, since (i) it essentially represents a challenging scenario; and (ii) all algorithms are evaluated on the same data. In particular, here, we employed 10 checkerboard (indoor) sequences and 12 cars-and-people (outdoor) sequences from the well-known Hopkins155 dataset~\cite{tron2007benchmark}. Moreover, we used another 8 outdoor sequences from the more recent MTPV dataset~\cite{li2013perspective}. To test the robustness of the different methods, we added different levels of Gaussian noise (with variance $\sigma^2 = $ 0.01, 0.02, 0.03, or 0.04)\footnote{Note that the intensities of the images are normalized to $[0,1]$. So the Gaussian noise with $\sigma^2=0.04$ is already big noise and more noise may never occur in practice.}  to the images. Altogether, this results in 150 evaluation sequences. The values of the parameters ($\gamma = 1.8\times 10^4$ and $\lambda = 1.0\times 10^4$) were tuned on a separate validation set and kept unchanged for all our experiments.


To compare the algorithms, we measure the number of tracking errors, \ie, the number of points that drift from the ground-truth by more than a certain error tolerance $\varepsilon$. Note that, in the sequences that we use, the ground-truth was obtained by the standard KLT tracker and then manually cleaned up, so the ground-truth itself contains some noise whose level depends on the scene itself. In particular, we observed that the ground-truth of the indoor checkerboard sequences generally has more noise than that of the outdoor sequences. Therefore, we set a larger error tolerance for the checkerboard sequences ($\varepsilon = 10$) than for the outdoor ones ($\varepsilon = 5$). For every sequence, we compute the average number of incorrectly tracked feature points over all the frames, and then average this number over the sequences.

\subsection{Hopkins Checkerboard Sequences}

\begin{table}[t!]
\footnotesize
\centering
\caption{Average number of tracking errors ($\varepsilon=10$) on the Hopkins checkerboard sequences with noise of different variances $\sigma^2$. The lower, the better.}
\vspace{-0.1cm}
\label{tab:avg_hopkins_10}
\begin{tabular}{  |  l  |   c   |   c   |   c   |      c       |   }
\hlinewd{0.5pt}\hlinewd{0.5pt}
                Methods &  KLT  & L1-KLT&  BFT  &{\bf Ours} \\
                \hline\hline
      $\sigma^2 = 0.00$ &  47.63& 34.69& 39.68 & \textbf{27.77}\\
      $\sigma^2 = 0.01$ &  46.92& 30.86& 39.30 & \textbf{27.32}\\
      $\sigma^2 = 0.02$ &  45.95& 29.69& 38.84 & \textbf{27.13}\\
      $\sigma^2 = 0.03$ &  46.59& 30.16& 39.16 & \textbf{28.18}\\
      $\sigma^2 = 0.04$ &  47.19& 31.16& 39.35 & \textbf{27.21}\\
\hlinewd{0.5pt}\hlinewd{0.5pt}
\end{tabular}
\vspace{-0.2cm}
\end{table}

We first evaluated our method and the baselines on the Hopkins checkerboard sequences, which depict controlled indoor scenes with multiple rigidly moving objects. The average number of tracks in this dataset is 202.9 .Generally, the repetitive texture in these sequences makes feature tracking more ambiguous and thus harder. However, in this experiment, we show that our multi-body feature tracker is more robust to this ambiguity. To provide a fair comparison, we used the same patch size ($7 \times 7$) and the same number of image pyramid levels (4) for all the methods. Furthermore, we initialized all the tracking methods with the ground-truth locations of the feature points in the first frame.

From Table~\ref{tab:avg_hopkins_10}, we can see that the L1-KLT tracker consistently achieves better results than the original KLT tracker and than BFT. Our algorithm, however, consistently outperforms L1-KLT, which clearly evidences the benefits of incorporating our multi-body prior. We observed that BFT generally fails to track moving objects, as illustrated in Fig.\ref{fig:hopkins_checkerboard}. This is mainly because BFT heavily relies on a good estimate of the global motion, obtained by registering the entire current image to the previous one. For scenes with multiple motions, however, global motion estimation becomes unreliable, thus causing BFT to fail to track the moving objects. Note that the performance of all the trackers remain relatively unaffected as the noise level increases. This is mainly due to the fact that the corners in the checkerboard, while resembling each other, are very strong features that are robust to noise. 

\subsection{Hopkins Car-and-People Sequences}

\begin{table}[t!]
\footnotesize
\centering
\caption{Average number of tracking errors ($\varepsilon=5$) on the Hopkins Car-and-People sequences with noise of different variances $\sigma^2$. The lower, the better.}
\vspace{-0.1cm}
\label{tab:avg_hopkins_12}
\begin{tabular}{  |  l  |   c   |   c   |   c   |      c       |   }
\hlinewd{0.5pt}\hlinewd{0.5pt}
                Methods &  KLT  & L1-KLT& BFT   &{\bf Ours} \\
                \hline\hline
      $\sigma^2 = 0.00$ &  21.71& 24.28 & 49.13 & \textbf{16.14}\\
      $\sigma^2 = 0.01$ &  34.59& 29.31 & 51.69 & \textbf{18.82}\\
      $\sigma^2 = 0.02$ &  54.95& 36.32 & 54.63 & \textbf{26.56}\\
      $\sigma^2 = 0.03$ &  76.02& 46.49 & 57.57 & \textbf{33.80}\\
      $\sigma^2 = 0.04$ &  95.17& 56.92 & 58.36 & \textbf{42.43}\\
\hlinewd{0.5pt}\hlinewd{0.5pt}
\end{tabular}
\vspace{-0.2cm}
\end{table}

We then evaluated the algorithms on the Hopkins Car-and-People sequences, depicting real-world outdoor scenes with multiple rigid motions. The number of tracks provided by the ground-truth ranges from 147 to 548 with an average of 369. Here, for all the methods, we used the same patch size and image pyramid levels as in the previous experiment, and initialized the feature points with their ground-truth locations in the first frame. The average number of tracking errors for the different methods under different image noise level is reported in Table~\ref{tab:avg_hopkins_12}. 
Again, our multi-body feature tracker achieves the lowest tracking error compared to the baselines, which confirms the robustness of our method.

\subsection{MTPV Sequences}

\begin{figure}[!t]
\centering
  \begin{tabular}{ccccc}
\hspace{-0.3cm}  \includegraphics[height=0.35\linewidth]{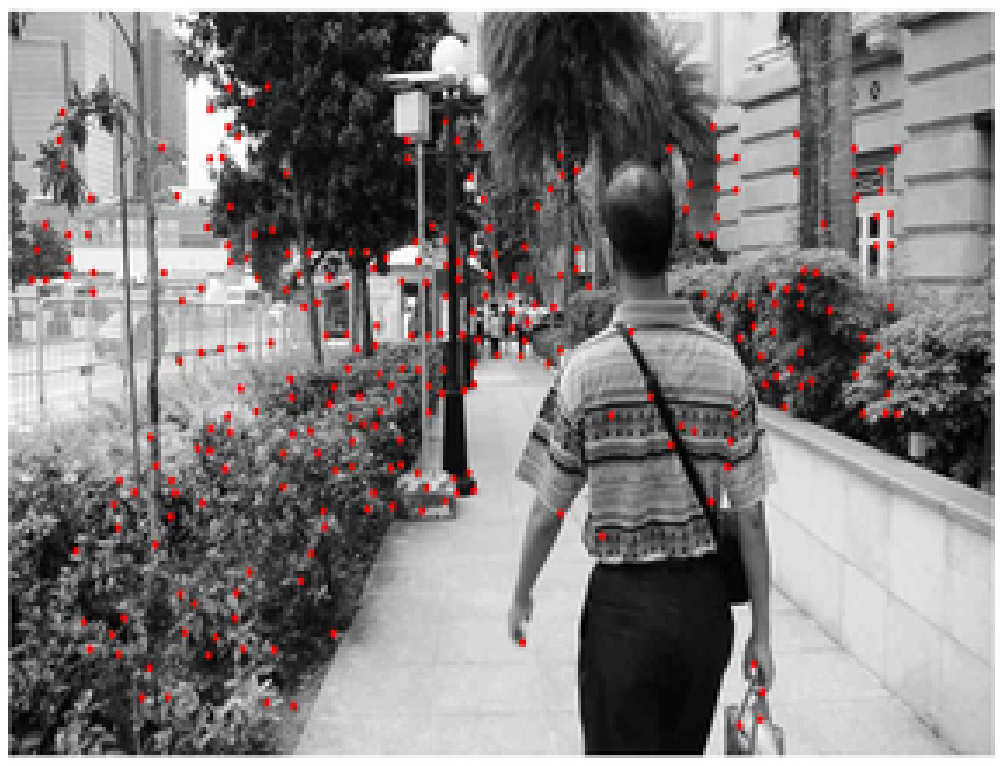} &
\hspace{-0.5cm}  \includegraphics[height=0.35\linewidth]{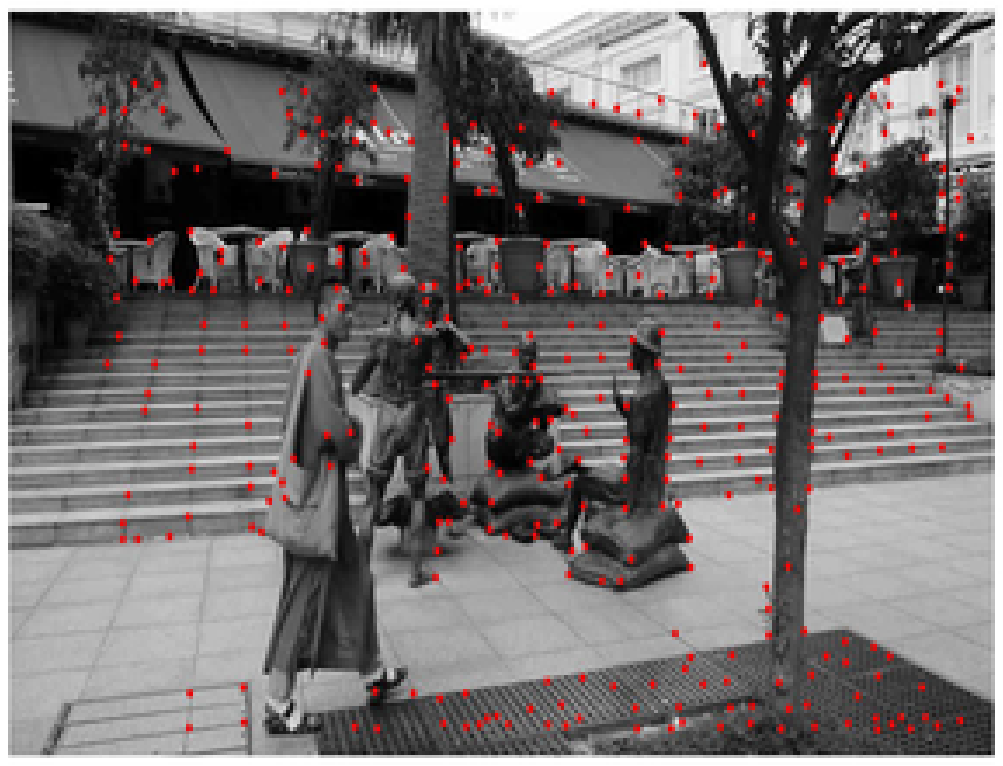}
\\
  \end{tabular}
  \vspace{-0.3cm}
  \caption{{\bf The MAN and MONK sequences of the MTPV dataset:} The feature points are marked as red. Note that the number of points on the walking man and monk is much smaller than on the background.}
  \label{fig:hopkins_man_monk}
\end{figure}

We further tested our method on the MTPV sequences, which provide images of higher quality and resolution\footnote{Note, however, that they are still highly compressed and not well-suited for tracking, as pointed out in the readme file of the dataset.} than the Hopkins dataset and contains sequences with strong perspective effects. By contrast, however, this dataset contains some outliers and missing data. For evaluation purpose, \ie, to create a complete and accurate ground-truth, we discarded the outliers and missing data. Since the image resolution is higher in this dataset, we used a larger patch size of $13 \times 13$ for all the methods. The results of all the algorithms are provided in Table~\ref{tab:avg_mptv_8}. Note that we still outperform all the baselines for most noise levels, with the exception of BFT for $\sigma^2 = 0.04$. We believe that the slightly less impressive gap between our approach and the baselines, in particular BFT, is due to the fact that the feature points in this dataset are often dominated by the background. See Fig.~\ref{fig:hopkins_man_monk} for typical examples of this dataset.

\begin{table}[h]
\footnotesize
\centering
\caption{Average number of tracking errors ($\varepsilon=5$) on the MTPV sequences with noise of different variances $\sigma^2$. The lower, the better.}
\vspace{-0.1cm}
\label{tab:avg_mptv_8}
\begin{tabular}{  |  l  |   c   |   c   |   c   |      c       |   }
\hlinewd{0.5pt}\hlinewd{0.5pt}
                Methods &  KLT  & L1-KLT&   BFT &{\bf Ours} \\
                \hline\hline
      $\sigma^2 = 0.00$ &  3.07& 13.34 &   6.83& \textbf{2.34}\\
      $\sigma^2 = 0.01$ & 17.76& 22.12 &   8.84& \textbf{3.87}\\
      $\sigma^2 = 0.02$ & 28.39& 27.26 &  11.17& \textbf{6.94}\\
      $\sigma^2 = 0.03$ & 40.61& 35.53 & 11.26 & \textbf{9.92}\\
      $\sigma^2 = 0.04$ & 47.69& 38.93 & \textbf{12.34} & 13.22\\
\hlinewd{0.5pt}\hlinewd{0.5pt}
\end{tabular}
\vspace{-0.2cm}
\end{table}

\subsection{KITTI Sequence}

\begin{figure*}[!t]
\centering
  \begin{tabular}{ccccc}
\hspace{-0.3cm}  \includegraphics[height=0.14\linewidth]{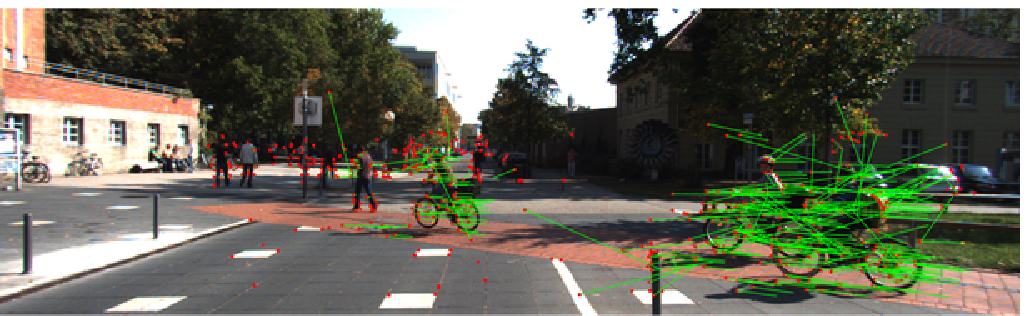} &
\hspace{-0.1cm}  \includegraphics[height=0.14\linewidth]{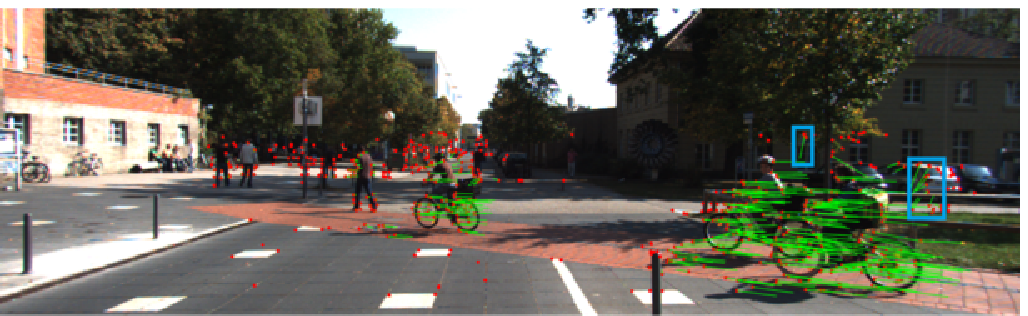} \\
\hspace{-0.3cm} KLT & \hspace{-0.1cm}L1-KLT \\
\hspace{-0.3cm}  \includegraphics[height=0.14\linewidth]{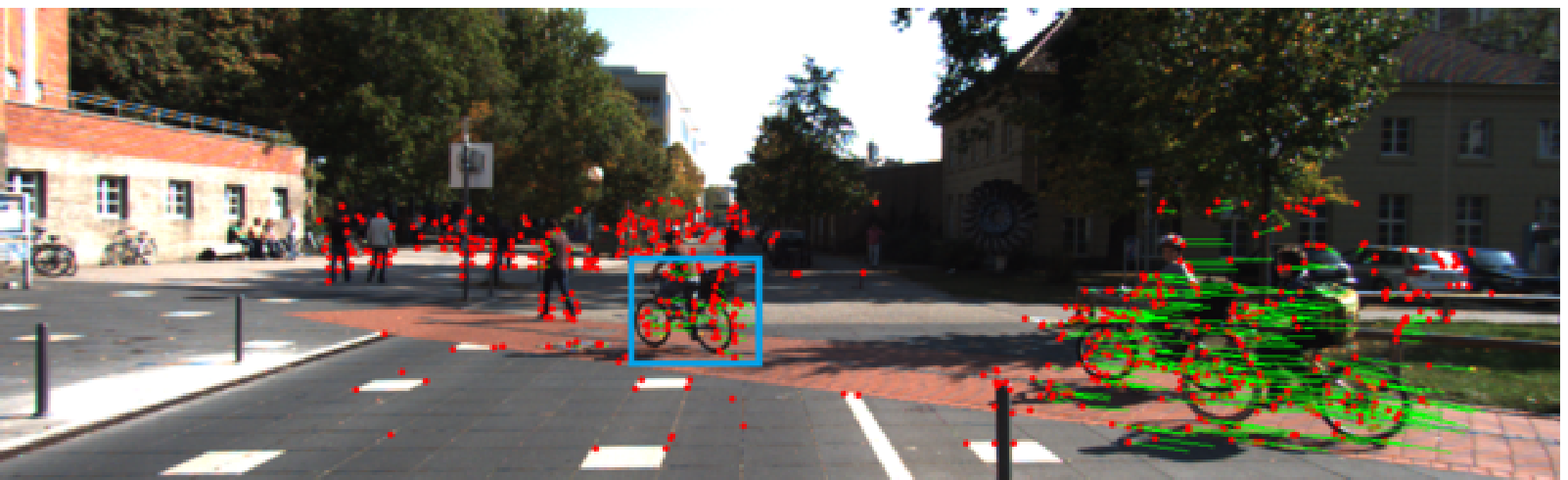} &
\hspace{-0.1cm}  \includegraphics[height=0.14\linewidth]{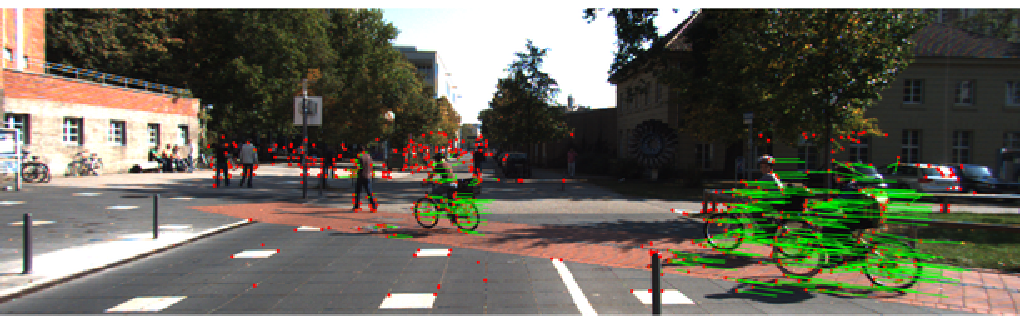}
\\
  \hspace{-0.3cm}BFT & \hspace{-0.1cm}Our Method
  \end{tabular}
  \vspace{-0.3cm}
  \caption{{\bf Performance of different trackers on the KITTI sequence:} The red points denote the current positions of the feature points, and the green lines the motion since the previous frame. As evidenced by the regions highlighted with a blue rectangle, L1-KLT and BFT make more tracking errors than our approach. Best viewed on screen with zoom-in.}
  \label{fig:KITTI}
\end{figure*}

To evaluate the algorithms on realistic, high-quality images, we employed four sequences\footnote{2011\_09\_26\_drive\_0018, 2011\_09\_26\_drive\_0051, 2011\_09\_26\_drive\_ 0056, and 2011\_09\_28\_drive\_0016.} from KITTI~\cite{Geiger2012CVPR}, depicting a street/traffic scene with multiple motions. Since no ground-truth trajectories are provided with this data, to obtain quantitative results,
we took 10 consecutive frames from each sequence, applied the KLT tracker to them, and manually cleaned up the results to get ground-truth trajectories with an average 177 points per sequence. The results of this experiments for different levels of noise added to the input are reported in Table~\ref{tab:avg_kitti}, and Fig.~\ref{fig:KITTI} shows a qualitative comparison of the algorithms. Note that our method also outperforms the baselines on this data. 

\begin{table}[h]
\footnotesize
\centering
\caption{Average number of tracking errors ($\varepsilon=5$) on the KITTI sequences with different noise variances $\sigma^2$. The lower, the better.}
\label{tab:avg_kitti}
\begin{tabular}{  |  l  |   c   |   c   |   c   |      c       |   }
\hline
                Methods &  KLT  & L1-KLT&   BFT &{\bf Ours} \\
                \hline\hline
      $\sigma^2 = 0.01$ & 21.43 & 22.05 & 27.48 & {\bf 14.18}   \\
      $\sigma^2 = 0.02$ & 24.35 & 22.85 & 27.80 & {\bf 16.70}   \\
      $\sigma^2 = 0.03$ & 31.15 & 26.88 & 27.85 & {\bf 17.70}   \\
      $\sigma^2 = 0.04$ & 34.43 & 29.23 & 27.75 & {\bf 20.33}   \\
\hline
\end{tabular}
\end{table}

\subsection{Frame-by-Frame Motion Segmentation}

\begin{table}[t!]
\scriptsize
\centering
\caption{Average error rate (in $\%$) of two-frame motion segmentation on the 22 Hopkins sequences with noise of different variances $\sigma^2$. The lower, the better.}
\vspace{-0.1cm}
\label{tab:avg_moseg2}
\begin{tabular}{  |  l  |   c   |   c   |   c   |   c   |      c       |   }
\hlinewd{0.5pt}\hlinewd{0.5pt}
      Methods &  KLT+SSC & KLT+EDSC & L1+SSC & L1+EDSC & {\bf Ours} \\
                \hline\hline
      $\sigma^2 = 0.00$ & 19.76  & 20.57 & 18.71 & 19.11 & \textbf{8.97}\\
      $\sigma^2 = 0.01$ & 19.76  & 20.61 & 19.61 & 20.41 & \textbf{9.35}\\
      $\sigma^2 = 0.02$ & 19.21  & 20.99 & 21.02 & 21.92 & \textbf{9.33}\\
      $\sigma^2 = 0.03$ & 20.63  & 20.69 & 22.21 & 20.48 & \textbf{9.89}\\
      $\sigma^2 = 0.04$ & 20.38  & 19.82 & 21.35 & 20.80 & \textbf{11.26}\\
\hlinewd{0.5pt}\hlinewd{0.5pt}
\end{tabular}
\vspace{-0.2cm}
\end{table}

In our formulation, we optimize our energy function w.r.t. two variables: the displacement vector ${\bf u}$ and the self-expressiveness coefficients ${\bf C}$. While the vector ${\bf u}$ provides the tracking results, the matrix ${\bf C}$, as in the subspace clustering literature, can be used to build an affinity matrix for spectral clustering, and thus, if we assume that the number of motions is known {\it a priori}, lets us perform motion segmentation. In other words, our method can also be interpreted as simultaneous feature tracking and frame-by-frame motion segmentation. In this experiment, we therefore aim to evaluate the frame-by-frame motion segmentation accuracy of our method.
Since, To the best of our knowledge, no existing motion segmentation methods perform feature tracking and frame-by-frame motion segmentation jointly, we compare our results with the following two-steps baselines: first, we find the tracks by KLT or L1-KLT and form the epipolar subspaces as in Eq.~\ref{eq: w_i}; second, we apply a subspace clustering method, \ie, Sparse  Subspace Clustering (SSC) or Efficient Dense Subspace Clustering (EDSC), to perform motion segmentation. This results in four baselines denoted by KLT+SSC, KLT+EDSC, L1+SSC~\cite{li2013perspective} and L1+EDSC. The results of motion segmentation on the 22 Hopkins sequences used previously are shown in Table~\ref{tab:avg_moseg2}. These results clearly evidence that our method outperforms the baselines significantly in terms of motion segmentation.

\section{Conclusion and Future Work}

In this paper, we have introduced a novel feature tracker that incorporates a multi-body rigidity prior into feature tracking. To this end, we have derived epipolar subspace constraints that prevent us from having to compute fundamental matrices and motion assignments explicitly. Our formulation only involves a series of convex subproblems, all of which have closed-from solutions. We have demonstrated the effectiveness of our method via extensive experiments on indoor and outdoor sequences. 

While adding global rigidity constraints (be it the low-rank or the epipolar subspace constraints) to the local KLT tracker improves robustness, it comes with some computational overhead. The current Matlab implementation of our method runs at about 1 frame per second for 200 points on a single core CPU (3.4GHZ), which is on par with BFT~\cite{poling2014better}, but slower than the original KLT tracker. In the future, we will therefore study how to speed up our approach, for instance by exploiting the GPU. Furthermore, our current model assumes that each patch undergoes only translation between consecutive frames. We therefore plan to investigate the use of more accurate models, such as affine transformations.

\section*{Appendix: ADMM Derivations}

Given the augmented Lagrangian in Eq.~\ref{eq:augm}, the ADMM subproblems can be derived as follows:

\noindent {\bf (1) } Computing ${\bf Z}$ can be expressed as the convex program
\begin{equation}
\label{eq:updateZ}
\min\limits_{\bf Z} \frac{\gamma}{\rho} \|{\bf Z}\|_1 + \frac{1}{2}\|{\bf Z} - ({\bf A}_{(\bf u)} - {\bf Y}_2/\rho)\|_F^2\;,
\end{equation}
which can be solved in closed-form by element-wise thresholding~\cite{cai2010singular}, which directly yields Eq.~\ref{eq:sol_Z}.

\noindent {\bf (2)} Similarly, computing ${\bf E}$ translates to
\begin{equation}
\min\limits_{\bf E}\frac{\lambda}{\rho} \|{\bf E}\|_1 + \frac{1}{2}\|{\bf E} - ({\bf W}_{(\bf m)}-{\bf W}_{(\bf m)}{\bf C} + {\bf Y}_1/\rho)\|_F^2\;,
\end{equation}
which again can be solved by element-wise thresholding, thus yielding Eq.~\ref{eq:sol_E}.

\noindent {\bf (3)} To compute ${\bf C}$, we have the least-squares problem
\begin{equation}
\min\limits_{\bf C} \frac{1}{2}\|{\bf C}\|_F^2 + \frac{\rho}{2}\|{\bf W}_{(\bf m)}{\bf C}- {\bf W}_{(\bf m)}+{\bf E}-{\bf Y}_1/\rho\|_F^2\;,
\end{equation}
which can easily be solved in closed-form as in Eq.~\ref{eq:sol_C}.

\noindent {\bf (4)} Computing ${\bf u}$ requires solving the problem
\begin{equation}
\begin{split}
\min\limits_{\bf u} \frac{\rho}{2}\|{\bf P}{\bf u}-{\bf m} - {\bf y}/\rho\|^2_2 -{\bf g}^T {\bf u} + \frac{\rho}{2}{\bf u}^T{\bf H}{\bf u}\;,
\end{split}
\end{equation}
where ${\bf g}$ is a column vector defined as
{\small
\begin{equation*}
{\bf g} = \bigg(\cdots,\sum\limits_{j}\big({\bf Y}_{2ij} +\rho(\tau_{ij}+{\bf Z}_{ij})\big)\triangledown I_{ij},\cdots\bigg)^T \in\mathbb{R}^{2N}\;,
\end{equation*}}
and ${\bf H}$ is a sparse block-diagonal matrix expressed as
{\small
\begin{equation*}
\label{defH}
{\bf H} = \begin{bmatrix} \ddots& & \\ &\sum\limits_j\triangledown I_{ij}^T\triangledown I_{ij} & \\ & &\ddots \end{bmatrix}\in\mathbb{R}^{2N\times 2N}\;.
\end{equation*}} This subproblem has again a closed-form solution given by Eq.~\ref{eq:sol_u}.
Note that ${\bf P}$ and ${\bf H}$ are sparse matrices, so ${\bf u}$ can be computed efficiently by sparse matrix techniques.

\noindent {\bf (5)} While solving for ${\bf m}$ may not seem straightforward, we show below that it is nothing but a least-squares problem. The subproblem w.r.t. ${\bf m}$ can be written as
\begin{equation}
\min\limits_{\bf m} \frac{\lambda}{2}\|{\bf W}_{(\bf m)}-{\bf W}_{(\bf m)}{\bf C}\|_F^2 + \frac{\rho}{2}\|{\bf m} - {\bf P}{\bf u}+{\bf y}/\rho\|_2^2\;.
\label{eq:sub_m}
\end{equation}
Let ${\bf M}$, ${\bf B}$, ${\bf G}$ $\in\mathbb{R}^{9\times N}$ be the matrix forms of $\bf m$, ${\bf b}$, ${\bf y/\rho} - {\bf P}{\bf u}$, respectively. Then,~\eqref{eq:sub_m} can be equivalently written as
\begin{equation}
\min\limits_{\bf M} \frac{\lambda}{2}\|{\bf M} ({\bf I} - {\bf C}) + {\bf B} ({\bf I} - {\bf C})\|_F^2 + \frac{\rho}{2}\|{\bf M} + {\bf G}\|_F^2\;.
\end{equation}
This again leads to a closed-form solution for ${\bf M}$ given by Eq.~\ref{eq:sol_m},
 where ${\bf Q} = ({\bf I}-{\bf C})({\bf I}-{\bf C})^T$ and ${\bf T} = ({\bf Y}_1/\rho - {\bf E})({\bf I}-{\bf C}^T)$.

{\small
\bibliographystyle{ieee}
\bibliography{multibodybib}

\begin{thebibliography}{10}\itemsep=-1pt

\bibitem{bouguet2001pyramidal}
J.-Y. Bouguet.
\newblock Pyramidal implementation of the affine lucas kanade feature tracker
  description of the algorithm.
\newblock {\em Technical Report, Intel Microprocessor Research Labs}, 2001.

\bibitem{boyd2011distributed}
S.~Boyd, N.~Parikh, E.~Chu, B.~Peleato, and J.~Eckstein.
\newblock Distributed optimization and statistical learning via the alternating
  direction method of multipliers.
\newblock {\em Foundations and Trends{\textregistered} in Machine Learning},
  3(1):1--122, 2011.

\bibitem{buchanan2007combining}
A.~Buchanan and A.~Fitzgibbon.
\newblock Combining local and global motion models for feature point tracking.
\newblock In {\em CVPR}, 2007.

\bibitem{cai2010singular}
J.-F. Cai, E.~J. Cand{\`e}s, and Z.~Shen.
\newblock A singular value thresholding algorithm for matrix completion.
\newblock {\em SIAM Journal on Optimization}, 20(4):1956--1982, 2010.

\bibitem{costeira1998multibody}
J.~P. Costeira and T.~Kanade.
\newblock A multibody factorization method for independently moving objects.
\newblock {\em IJCV}, 29(3):159--179, 1998.

\bibitem{elhamifar2013sparse}
E.~Elhamifar and R.~Vidal.
\newblock Sparse subspace clustering: Algorithm, theory, and applications.
\newblock {\em PAMI}, 35(11):2765--2781, 2013.

\bibitem{garg2011dense}
R.~Garg, L.~Pizarro, D.~Rueckert, and L.~Agapito.
\newblock Dense multi-frame optic flow for non-rigid objects using subspace
  constraints.
\newblock In {\em ACCV}, 2010.

\bibitem{garg2013variational}
R.~Garg, A.~Roussos, and L.~Agapito.
\newblock A variational approach to video registration with subspace
  constraints.
\newblock {\em IJCV}, 104(3):286--314, 2013.

\bibitem{Geiger2012CVPR}
A.~Geiger, P.~Lenz, and R.~Urtasun.
\newblock Are we ready for autonomous driving? the kitti vision benchmark
  suite.
\newblock In {\em CVPR}, 2012.

\bibitem{Hartley2004}
R.~I. Hartley and A.~Zisserman.
\newblock {\em Multiple View Geometry in Computer Vision}.
\newblock Cambridge University Press, ISBN: 0521540518, second edition, 2004.

\bibitem{ji2014robust}
P.~Ji, H.~Li, M.~Salzmann, and Y.~Dai.
\newblock Robust motion segmentation with unknown correspondences.
\newblock In {\em ECCV}. 2014.

\bibitem{ji2014efficient}
P.~Ji, M.~Salzmann, and H.~Li.
\newblock Efficient dense subspace clustering.
\newblock In {\em WACV}, 2014.

\bibitem{ji2015shape}
P.~Ji, M.~Salzmann, and H.~Li.
\newblock Shape interaction matrix revisited and robustified: Efficient
  subspace clustering with corrupted and incomplete data.
\newblock In {\em ICCV}, 2015.

\bibitem{ji2014null}
P.~Ji, Y.~Zhong, H.~Li, and M.~Salzmann.
\newblock Null space clustering with applications to motion segmentation and
  face clustering.
\newblock In {\em ICIP}, 2014.

\bibitem{li2007two}
H.~Li.
\newblock Two-view motion segmentation from linear programming relaxation.
\newblock In {\em CVPR}, 2007.

\bibitem{li2013perspective}
Z.~Li, J.~Guo, L.-F. Cheong, and S.~Z. Zhou.
\newblock Perspective motion segmentation via collaborative clustering.
\newblock In {\em ICCV}, 2013.

\bibitem{liu2013robust}
G.~Liu, Z.~Lin, S.~Yan, J.~Sun, Y.~Yu, and Y.~Ma.
\newblock Robust recovery of subspace structures by low-rank representation.
\newblock {\em PAMI}, 35(1):171--184, 2013.

\bibitem{lucas1981iterative}
B.~D. Lucas, T.~Kanade, et~al.
\newblock An iterative image registration technique with an application to
  stereo vision.
\newblock In {\em IJCAI}, volume~81, pages 674--679, 1981.

\bibitem{piccini2014good}
T.~Piccini, M.~Persson, K.~Nordberg, M.~Felsberg, and R.~Mester.
\newblock Good edgels to track: Beating the aperture problem with epipolar
  geometry.
\newblock In {\em ECCV Workshops}, 2014.

\bibitem{poling2014better}
B.~Poling, G.~Lerman, and A.~Szlam.
\newblock Better feature tracking through subspace constraints.
\newblock In {\em CVPR}, 2014.

\bibitem{shi1994good}
J.~Shi and C.~Tomasi.
\newblock Good features to track.
\newblock In {\em CVPR}, 1994.

\bibitem{szeliski2010computer}
R.~Szeliski.
\newblock {\em Computer vision: algorithms and applications}.
\newblock Springer Science \& Business Media, 2010.

\bibitem{tomasi1991detection}
C.~Tomasi and T.~Kanade.
\newblock {\em Detection and tracking of point features}.
\newblock School of Computer Science, Carnegie Mellon Univ. Pittsburgh, 1991.

\bibitem{torresani2002space}
L.~Torresani and C.~Bregler.
\newblock Space-time tracking.
\newblock In {\em ECCV}, 2002.

\bibitem{tron2007benchmark}
R.~Tron and R.~Vidal.
\newblock A benchmark for the comparison of 3-d motion segmentation algorithms.
\newblock In {\em CVPR}, 2007.

\bibitem{valgaerts2008variational}
L.~Valgaerts, A.~Bruhn, and J.~Weickert.
\newblock A variational model for the joint recovery of the fundamental matrix
  and the optical flow.
\newblock In {\em Pattern Recognition}, pages 314--324. 2008.

\bibitem{vidal2005generalized}
R.~Vidal, Y.~Ma, and S.~Sastry.
\newblock Generalized principal component analysis ({GPCA}).
\newblock {\em PAMI}, 27(12):1945--1959, 2005.

\bibitem{vidal2002segmentation}
R.~Vidal, S.~Soatto, Y.~Ma, and S.~Sastry.
\newblock Segmentation of dynamic scenes from the multibody fundamental matrix.
\newblock In {\em CVPR}, 2001.

\bibitem{wedel2009structure}
A.~Wedel, D.~Cremers, T.~Pock, and H.~Bischof.
\newblock Structure-and motion-adaptive regularization for high accuracy optic
  flow.
\newblock In {\em ICCV}, 2009.

\bibitem{wedel2008duality}
A.~Wedel, T.~Pock, J.~Braun, U.~Franke, and D.~Cremers.
\newblock Duality {TV-L}1 flow with fundamental matrix prior.
\newblock In {\em IVCNZ}, 2008.

\bibitem{yan2006general}
J.~Yan and M.~Pollefeys.
\newblock A general framework for motion segmentation: Independent,
  articulated, rigid, non-rigid, degenerate and non-degenerate.
\newblock In {\em ECCV}, 2006.

\end{thebibliography}



@inproceedings{tron2007benchmark,
  title={A benchmark for the comparison of 3-d motion segmentation algorithms},
  author={Tron, Roberto and Vidal, Ren{\'e}},
  booktitle={CVPR},
  year={2007}
}

@inproceedings{li2013perspective,
  title={Perspective motion segmentation via collaborative clustering},
  author={Li, Zhuwen and Guo, Jiaming and Cheong, Loong-Fah and Zhou, Steven Zhiying},
  booktitle={ICCV},
  year={2013}
}

@inproceedings{vidal2002segmentation,
  title={Segmentation of dynamic scenes from the multibody fundamental matrix},
  author={Vidal, Ren{\'e} and Soatto, Stefano and Ma, Yi and Sastry, Shankar},
  booktitle={CVPR},
  year={2001}
}

@article{costeira1998multibody,
  title={A multibody factorization method for independently moving objects},
  author={Costeira, Jo{\~a}o Paulo and Kanade, Takeo},
  journal={IJCV},
  volume={29},
  number={3},
  pages={159--179},
  year={1998}
}

@inproceedings{li2007two,
  title={Two-view motion segmentation from linear programming relaxation},
  author={Li, Hongdong},
  booktitle={CVPR},
  year={2007}
}

@inproceedings{jung2014rigid,
  title={Rigid Motion Segmentation using Randomized Voting},
  author={Jung, Heechul and Ju, Jeongwoo and Kim, Junmo},
  booktitle={CVPR},
  year={2014}
}

@inproceedings{poling2014better,
  title={Better Feature Tracking Through Subspace Constraints},
  author={Poling, Bryan and Lerman, Gilad and Szlam, Arthur},
  booktitle={CVPR},
  year={2014}
}

@article{baker2004lucas,
  title={Lucas-kanade 20 years on: A unifying framework},
  author={Baker, Simon and Matthews, Iain},
  journal={IJCV},
  volume={56},
  number={3},
  pages={221--255},
  year={2004}
}

@inproceedings{lucas1981iterative,
  title={An iterative image registration technique with an application to stereo vision.},
  author={Lucas, Bruce D and Kanade, Takeo and others},
  booktitle={IJCAI},
  volume={81},
  pages={674--679},
  year={1981}
}

@inproceedings{shi1994good,
  title={Good features to track},
  author={Shi, Jianbo and Tomasi, Carlo},
  booktitle={CVPR},
  year={1994}
}

@book{tomasi1991detection,
  title={Detection and tracking of point features},
  author={Tomasi, Carlo and Kanade, Takeo},
  year={1991},
  booktitle={Technical Report CMU-CS-91-132},
  publisher={School of Computer Science, Carnegie Mellon Univ. Pittsburgh}
}

@article{elhamifar2013sparse,
  title={Sparse subspace clustering: Algorithm, theory, and applications},
  author={Elhamifar, Ehsan and Vidal, Ren{\'e}},
  journal={PAMI},
  volume={35},
  number={11},
  pages={2765--2781},
  year={2013}
}

@inproceedings{ji2014efficient,
  title={Efficient dense subspace clustering},
  author={Ji, Pan and Salzmann, Mathieu and Li, Hongdong},
  booktitle={WACV},
  year={2014}
}

@inproceedings{buchanan2007combining,
  title={Combining local and global motion models for feature point tracking},
  author={Buchanan, Aeron and Fitzgibbon, Andrew},
  booktitle={CVPR},
  year={2007}
}

@article{garg2013variational,
  title={A variational approach to video registration with subspace constraints},
  author={Garg, Ravi and Roussos, Anastasios and Agapito, Lourdes},
  journal={IJCV},
  volume={104},
  number={3},
  pages={286--314},
  year={2013}
}

@inproceedings{torresani2001tracking,
  title={Tracking and modeling non-rigid objects with rank constraints},
  author={Torresani, Lorenzo and Yang, Danny B and Alexander, Eugene J and Bregler, Christoph},
  booktitle={CVPR},
  year={2001}
}

@incollection{slesareva2005optic,
  title={Optic flow goes stereo: A variational method for estimating discontinuity-preserving dense disparity maps},
  author={Slesareva, Natalia and Bruhn, Andr{\'e}s and Weickert, Joachim},
  booktitle={DAGM},
  pages={33--40},
  year={2005}
}

@incollection{valgaerts2008variational,
  title={A variational model for the joint recovery of the fundamental matrix and the optical flow},
  author={Valgaerts, Levi and Bruhn, Andr{\'e}s and Weickert, Joachim},
  booktitle={Pattern Recognition},
  pages={314--324},
  year={2008}
}

@inproceedings{wedel2008duality,
  title={Duality {TV-L}1 flow with fundamental matrix prior},
  author={Wedel, A and Pock, T and Braun, J and Franke, U and Cremers, D},
  booktitle={IVCNZ},
  year={2008}
}

@inproceedings{piccini2014good,
  title={Good Edgels to Track: Beating the Aperture Problem with Epipolar Geometry},
  author={Piccini, Tommaso and Persson, Mikael and Nordberg, Klas and Felsberg, Michael and Mester, Rudolf},
  booktitle={ECCV Workshops},
  year={2014}
}

@inproceedings{torresani2002space,
  title={Space-time tracking},
  author={Torresani, Lorenzo and Bregler, Christoph},
  booktitle={ECCV},
  year={2002}
}

@article{bouguet2001pyramidal,
  title={Pyramidal implementation of the affine lucas kanade feature tracker description of the algorithm},
  author={Bouguet, Jean-Yves},
  journal={Technical Report, Intel Microprocessor Research Labs},
  year={2001}
}

@inproceedings{black1993framework,
  title={A framework for the robust estimation of optical flow},
  author={Black, Michael J and Anandan, P},
  booktitle={ICCV},
  year={1993}
}

@inproceedings{brox2004high,
  title={High accuracy optical flow estimation based on a theory for warping},
  author={Brox, Thomas and Bruhn, Andr{\'e}s and Papenberg, Nils and Weickert, Joachim},
  booktitle={ECCV},
  year={2004}
}

@Book{Hartley2004,
    author = "Hartley, R.~I. and Zisserman, A.",
    title = "Multiple View Geometry in Computer Vision",
    edition = "Second",
    year = "2004",
    publisher = "Cambridge University Press, ISBN: 0521540518"
}

@article{boyd2011distributed,
  title={Distributed optimization and statistical learning via the alternating direction method of multipliers},
  author={Boyd, Stephen and Parikh, Neal and Chu, Eric and Peleato, Borja and Eckstein, Jonathan},
  journal={Foundations and Trends{\textregistered} in Machine Learning},
  volume={3},
  number={1},
  pages={1--122},
  year={2011}
}

@article{cai2010singular,
  title={A singular value thresholding algorithm for matrix completion},
  author={Cai, Jian-Feng and Cand{\`e}s, Emmanuel J and Shen, Zuowei},
  journal={SIAM Journal on Optimization},
  volume={20},
  number={4},
  pages={1956--1982},
  year={2010}
}

@article{zhang1995robust,
  title={A robust technique for matching two uncalibrated images through the recovery of the unknown epipolar geometry},
  author={Zhang, Zhengyou and Deriche, Rachid and Faugeras, Olivier and Luong, Quang-Tuan},
  journal={Artificial intelligence},
  volume={78},
  number={1},
  pages={87--119},
  year={1995}
}

@article{liu2013robust,
  title={Robust recovery of subspace structures by low-rank representation},
  author={Liu, Guangcan and Lin, Zhouchen and Yan, Shuicheng and Sun, Ju and Yu, Yong and Ma, Yi},
  journal={PAMI},
  volume={35},
  number={1},
  pages={171--184},
  year={2013}
}

@article{vidal2005generalized,
  title={Generalized principal component analysis ({GPCA})},
  author={Vidal, Rene and Ma, Yi and Sastry, Shankar},
  journal={PAMI},
  volume={27},
  number={12},
  pages={1945--1959},
  year={2005}
}

@INPROCEEDINGS{ji2014null,
author={Pan Ji and Yiran Zhong and Hongdong Li and Salzmann, M.},
booktitle={ICIP},
title={Null space clustering with applications to motion segmentation and face clustering},
year={2014}
}

@inproceedings{yan2006general,
  title={A general framework for motion segmentation: Independent, articulated, rigid, non-rigid, degenerate and non-degenerate},
  author={Yan, Jingyu and Pollefeys, Marc},
  booktitle={ECCV},
  year={2006}
}

@article{brox2011large,
  title={Large displacement optical flow: descriptor matching in variational motion estimation},
  author={Brox, Thomas and Malik, Jitendra},
  journal={PAMI},
  volume={33},
  number={3},
  pages={500--513},
  year={2011}
}

@inproceedings{horn1981determining,
  title={Determining optical flow},
  author={Horn, Berthold K and Schunck, Brian G},
  booktitle={Technical symposium east},
  pages={319--331},
  year={1981},
  organization={International Society for Optics and Photonics}
}

@article{bruhn2005lucas,
  title={Lucas/Kanade meets Horn/Schunck: Combining local and global optic flow methods},
  author={Bruhn, Andr{\'e}s and Weickert, Joachim and Schn{\"o}rr, Christoph},
  journal={IJCV},
  volume={61},
  number={3},
  pages={211--231},
  year={2005}
}

@inproceedings{wedel2009structure,
  title={Structure-and motion-adaptive regularization for high accuracy optic flow},
  author={Wedel, Andreas and Cremers, Daniel and Pock, Thomas and Bischof, Horst},
  booktitle={ICCV},
  year={2009}
}

@inproceedings{garg2011dense,
  title={Dense multi-frame optic flow for non-rigid objects using subspace constraints},
  author={Garg, Ravi and Pizarro, Luis and Rueckert, Daniel and Agapito, Lourdes},
  booktitle={ACCV},
  year={2010}
}

@INPROCEEDINGS{Geiger2012CVPR,
  author = {Andreas Geiger and Philip Lenz and Raquel Urtasun},
  title = {Are we ready for Autonomous Driving? The KITTI Vision Benchmark Suite},
  booktitle = {CVPR},
  year = {2012}
}

@inproceedings{elhamifar2010clustering,
  title={Clustering disjoint subspaces via sparse representation},
  author={Elhamifar, Ehsan and Vidal, Ren{\'e}},
  booktitle={ICASSP},
  year={2010}
}
}

\end{document}